\documentclass[11pt]{article}

\usepackage{epsfig}
\usepackage{graphicx} 
\usepackage{amsmath}
\usepackage{amssymb, verbatim}
\usepackage{fullpage}
\usepackage{ctable}
\usepackage{multirow}

\begin{document}

\title{Fully Automatic Expression-Invariant Face Correspondence}
\author{Augusto Salazar\footnote{Perception and Intelligent Control research group, National University of Colombia.} $^\dag$ \and Stefanie Wuhrer\footnote{Cluster of Excellence, Multimodal Computing and Interaction, Saarland University.} \footnote{National Research Council of Canada.} \and Chang Shu$^\ddag$  \and Flavio Prieto \footnote{GAUNAL research group, National University of Colombia.}}

\maketitle

\begin{abstract}
We consider the problem of computing accurate point-to-point correspondences among a set of human face scans with varying expressions. Our fully automatic approach does not require any manually placed markers on the scan. Instead, the approach learns the locations of a set of landmarks present in a database and uses this knowledge to automatically predict the locations of these landmarks on a newly available scan. The predicted landmarks are then used to compute point-to-point correspondences between a template model and the newly available scan. To accurately fit the expression of the template to the expression of the scan, we use as template a blendshape model. Our algorithm was tested on a database of human faces of different ethnic groups with strongly varying expressions. Experimental results show  that the obtained point-to-point correspondence is both highly accurate and consistent for most of the tested 3D face models.
\end{abstract}

\section{Introduction}
\label{sec:intro}

We consider the problem of computing point-to-point correspondences among a set of human face scans with varying expressions in a fully automatic way. This problem arises from building a statistical model that encodes face shape and expression simultaneously using a database of human face scans. In order to build a statistical model, we rely on the correct computation of dense point-to-point correspondences among the subjects of a database. That is, the raw scans have to be parameterized in such a way that likewise anatomical parts correspond across the models~\cite{Dryden02}. Facial expression affects the geometry of the human face and therefore is important for facial shape analysis. A statistical model of face shapes and expressions can be used in applications such as face recognition, expression recognition, or reconstructing accurate 3D models of faces from input images~\cite{Blanz99,Jiang05,Romdhani02,Romdhani03,Brunton11}.

Computing accurate point-to-point correspondences for a set of face shapes in varying expressions is a challenging task because the face shape varies across the database and each subject has its own way to perform facial expressions. The problem is further complicated by incomplete and noisy data in the scans.

While many approaches have been proposed to compute point-to-point correspondences~\cite{vanKaick11}, only few of them have been applied to statistical model building and shape analysis of human face shapes. Blanz and Vetter~\cite{Blanz99} built a statistical model called morphable model for a set of 3D face scans with varying expressions. The correspondence algorithm is based on using optical flow on the texture information of the faces. This assumes that the faces are approximately spatially aligned. Xi and Shu~\cite{Xi09} built a statistical model based on principal component analysis for a set of 3D face scans with neutral expressions. The correspondence algorithm is based on fitting a template model to the scans using a non-rigid iterative closest point algorithm. To start this algorithm, the faces need to be approximately aligned using a set of manually placed marker positions. Both of these registration approaches fail for misaligned models.

In this work, we develop a novel technique to compute correspondences between a set of facial scans with varying expressions that does not require the scans to be spatially aligned. Our correspondence computation procedure uses a template model $P$ as prior knowledge on the geometry of the face shapes. Unlike Xi and Shu~\cite{Xi09}, we aim to find correspondences for faces with varying expressions. Hence, it is not enough to have a template model that captures the face shape of a generic model, but we also need to capture the expressions of a generic model. To achieve this, we model $P$ as a blendshape model as in Li et al.~\cite{Li10}. In a blendshape model, expressions are modeled as a linear combination of a set of basic expressions. Hence, blendshape models are both simple and effective to model facial expressions.

\begin{figure}[htp]
\centering
\includegraphics[width=\textwidth,height=!]{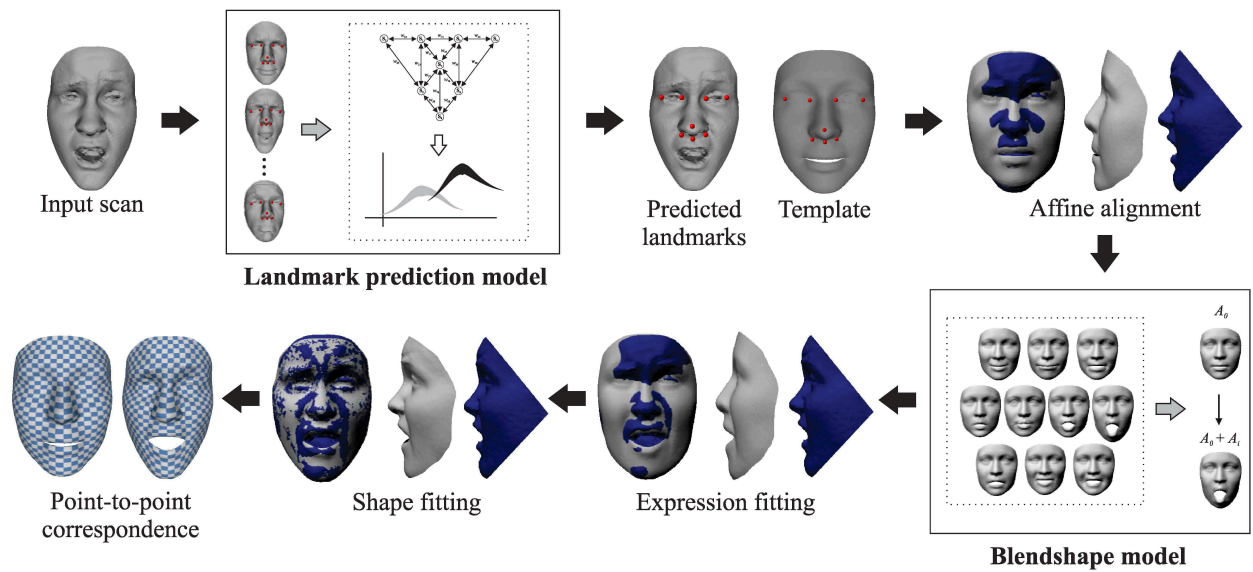}
\caption{Overview of the fully automatic expression-invariant face correspondence approach.}\label{fig:graphabstract}
\end{figure}

Our approach proceeds as follows. We first use a database of human face scans with manually placed landmark positions to learn local properties and spatial relationships between the landmarks using a Markov network. Given an input scan $F$ without manually placed landmarks, we first predict the landmark positions on $F$ by carrying out statistical inference over the trained Markov network. Sections \ref{sec:learning} and \ref{sec:prediction} discuss this step. In order to perform statistical inference, we need to restrict the search region for each landmark. This is detailed in Sections~\ref{restricting} to \ref{alignment}. The predicted landmarks are used to align $P$ to $F$. In order to fit the expression of $P$ to the expression of $F$, the template is aligned to the scan as outlined in Section~\ref{sec:affine} and the weights of the generic blendshape model are optimized as discussed in Section~\ref{sec:expression}. Finally, the shape of $P$ is changed to fit the shape of $F$ as outlined in Section ~\ref{sec:shape}. Fig.~\ref{fig:graphabstract} shows an overview of the method.

\section{Related Work}
\label{sec:related}

This section reviews literature in face shape analysis related to finding landmarks on face models, computing correspondences between three-dimensional shapes, and using blendshape models for facial animation.

\subsection{Finding Landmarks on Face Models}

Traditionally, facial feature detection is done in 2D images, but recent developments on 3D data acquisition have allowed to overcome the problems attached to the 3D technologies. Existing registration methods demonstrated that landmark-based methods provide more accurate and consistent results. However, only a few approaches consider 3D landmark detection, while accounting for expression and pose variations~\cite{Mehryar10}.

Ben Azouz et al.~\cite{azouz06} propose a method to find correspondences by automatically predicting marker positions on 3D models of a human body. The method encodes the statistics of a surface descriptor  and geometric properties at the locations of manually placed landmarks in a Markov network. This method works only for models with slight variation of posture. Mehryar et al.~\cite{Mehryar10} introduce an algorithm to automatically detect eyes, nose, and mouth on 3D faces. The algorithm correctly detects the landmarks in the presence of pose, facial expression and occlusion variations. This method is useful as initial alignment but not for an accurate registration. Berretti et al.~\cite{berretti11} combine principal curvatures analysis, edge detector and SIFT descriptors to find 9 landmarks on the eyes nose and mouth regions in range images. The landmarks are properly detected in the presence of facial expressions but the method relies on anthropometric facial proportions to define the search regions and assumes that the face is upright oriented. Creusot et al.~\cite{Creusot2010} present a method to localize a set of 13 facial landmark points under large pose variation or when occlusion is present. Their method learns the properties of a set of descriptors computed at the landmark locations and encodes both local information and spatial relationships into a graph. The method works well for neutral pose. However, in the presence of expression variation, the accuracy decreases considerably. Segundo et al.~\cite{segundo10} develop a method for face segmentation and landmark detection in range images. The landmark detection method combines surface curvature information and depth relief curve analysis to find 5 landmarks located on the nose and eye regions. The landmarks are properly detected in the presence of facial expressions and hair occlusions, but the method relies on a specific acquisition setup. Perakis et al.~\cite{perakis09,perakis10} present a method to detect landmarks under large pose variations using a Active Landmark Model (ALM), which is a statistical shape model learned from 8 manually annotated landmarks. Using a combination of the Shape Index descriptor and Spin Images, the search space for the fitting of the ALM is defined. The final set of landmarks is defined by selecting the set of candidates that satisfies the geometric restrictions encoded in the ALM. The experiments show that the method works in the presence of facial expressions and pose variation up to 80 degrees around the y-axis. Nair and Cavallaro~\cite{nair09} use a point distribution model to estimate the location of 49 landmarks on the eyebrow, eye and nose regions. The method works well in the presence of expressions and noisy data. However the error in the localization of landmarks is quite high (a comparison of the results is provided in Section~\ref{sec:accuracy}). Lu and Jain~\cite{lu06} present a multimodal approach for facial feature extraction. The nose tip is located using only the 3D information, and the eyes and mouth corners are extracted using 2D and 3D data. As their focus is handling changes in head pose and lighting conditions, variations due to facial expressions are not considered in their experiments. This multimodal approach is used by Lu et al.~\cite{lu06b} as part of a system for face recognition in the presence of pose and expression variation (only smiling expression variations are included in the test data). The authors claim that the expression changes decrease the accuracy of the system. However, quantitative results of the landmark detection are not provided. In addition, the requirement of the texture data is a limitation of the multimodal approaches because sometimes such information is not available.

As our aim is to obtain accurate point-to-point correspondences, we derived a landmark prediction method based on the approach of Ben Azouz et al.~\cite{azouz06}. The surface descriptor we used is able to catch the local geometry properly~\cite{sun01} and, by combining it with a canonical representation~\cite{elad03}, our approach is able to detect landmarks in the presence of facial expressions. We select a machine learning-based approach to avoid classic assumptions such as: the nose tip is the closest point to the camera~\cite{chang06}, the inner-corners of the eyes and the tip of the nose are the most salient points~\cite{segundo10}, the 3D face scan is in a frontal upright canonical pose~\cite{berretti11}, among others. The advantage is that learning-based approaches can easily be extended to other contexts.

\subsection{Correspondence Computation}

Several methods have been proposed to solve the problem of establishing a meaningful correspondence between shapes. Here, we focus on computing correspondences between human face shapes. Methods that do not assume templates usually have the problem that some points are not registered accurately. To remedy this, we assume a template model. In the following, we only review approaches that use template models (for details about methods for correspondence computation see the survey of van Kaick et al.~\cite{vanKaick11}).

Passalis et al.~\cite{passalis11} proposed a 3D face recognition method that uses facial symmetry to handle pose variation and missing data. A template is fitted to the shape of the input model as follows: an Annotated Face Model~\cite{kakadiaris07} is iteratively deformed towards the input using automatically predicted landmarks and an algorithm based on Simulated Annealing. When dealing with facial expressions, the performance of the recognition system decreases. This is due to an incorrect registration of the mouth region. The authors do not show extensive evaluations of this fully-automatic registration method as this is not the main part of their work.

Statistical learning-based approaches have been effectively used to model facial variations oriented to both the synthesis and recognition of faces. Blanz and Vetter~\cite{Blanz99} developed a 3D morphable model (3DMM) for the synthesis of 3D faces from photographs. As the registration is specific to the scanning setup, rigid alignment of the scans is assumed. Lu and Jain~\cite{Lu06a} present an approach to perform face recognition using 3D face scans. The approach  builds a 3DMM for each subject in the database. When a test image becomes available, the approach matches the scan to a specific individual using the learned 3DMM. Unlike our method, their training data is parameterized using manually placed landmarks and the test scans are parameterized using individual-specific deformation models. Basso et al.~\cite{basso06} extend the method of Blanz and Vetter~\cite{Blanz99} to register 3D scans of faces with arbitrary identity and expression. The rigid alignment of the scans is also assumed for registration. To avoid the use of texture information, Amberg et al.~\cite{amberg08} present a method to fit a 3DMM to 3D face scans using only shape information. They demonstrate the performance of the method in the presence of expression variation, occlusion and missing data, but do not conduct extensive evaluations of the registration. 

Registration methods based on iteratively deforming a template to the data are an alternative to statistical learning-based approaches. Allen et al.~\cite{Allen03} present an approach to parameterize a set of 3D scans of human body shapes in similar posture. To fit the template to each scan, the method proceeds by using a non-rigid iterative closest point (ICP) framework coupled with a set of manually placed marker positions. Xi and Shu~\cite{Xi09} extend the method of Allen et al.~\cite{Allen03} to deform a template model to a head scan. The shape fitting is carried out as in Allen et al.~\cite{Allen03} but uses radial basis functions to speed up the deformation process. Unlike our method, this only allows for neutral expressions and uses manually placed markers to align the template to a head scan. Wuhrer et al.~\cite{wuhrer2011a} propose a method to deform a template model to a human body scan in arbitrary posture. The method works in two stages: posture and shape fitting. Posture fitting relies on the location of different landmarks, which are predicted in a fully automatic way using a statistical model of landmark positions learned from a population. Our method can be viewed as an extension of this approach, but instead of fitting the posture, we fit the expression using blendshapes (see Section~\ref{sec:blenshapes}).

Methods that compute a correspondence between two surfaces by embedding the intrinsic geometry of one surface into the other one by using Generalized Multi-Dimensional Scaling (GMDS)~\cite{Bronstein06} are another alternative to deal with variations due to facial expressions~\cite{Bronstein07}. The performance of these methods has been demonstrated for face recognition. As GMDS methods do not take care that close-by points on one surface map to close-by points on the other, the results are often spatially inconsistent. This prevents such methods from being used for shape analysis.

\subsection{Use of Blendshape Models}\label{sec:blenshapes}

Modeling expressions using blendshape models is an alternative to approaches based on statistical models where a comprehensive database annotation process has to be carried out to extract variational information. In a blendshape model, movements of the different facial regions are assumed to be independent. Any expression is then modeled as a linear combination of the differences between a set of basic expressions, called \textit{blendshapes}, and a neutral expression. That is, to produce an expression, the displacements causing the movement are linearly combined. Using a representative set of blendshapes, this simple model is effective to model facial expressions.

Li et al.~\cite{Li10} propose a method to transfer the expression of a subject to an animated character. Their framework allows to create optimal blendshapes from a set of example poses of a digital face model automatically. Weise et al.~\cite{Weise11} present a framework for real-time 3D facial animation. The method tracks the rigid and non-rigid motion of the user's face accurately. They incorporate the expression transfer approach of Li et al.~\cite{Li10} in order to find much of the variation from the example expressions. The registration stage requires offline training where a generic template is fitted to the face of a specific subject. To obtain the results, manual marking of features has to be carried out.

Because of the advantages of modeling expression using linear blend\-shapes, we use it to aid the shape matching. We only optimize a blending weight per expression. This reduces the dimensionality of the optimization space drastically. Since our database of blendshapes is small, the expression fitting stage of our algorithm is efficient and helps to improve the results significantly.

\section{Landmark Prediction}
\label{sec:landmark}

\begin{figure}[htp]
\centering
\includegraphics[width=0.5\textwidth,height=!]{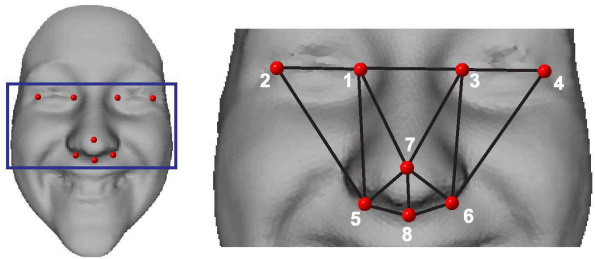}
\caption{Face model with landmarks. Locations and landmark graph structure.}\label{fig:landmarks}
\end{figure}

This section outlines how to predict a set of landmark positions on a face scan. To establish the correspondences across the whole database, we fit a template to each model. The fitting process begins with the extraction of the locations of eight landmarks shown as red spheres in Fig. \ref{fig:landmarks}. The locations of the landmarks were selected based on the fact that in the presence of facial expressions, the corners of the eyes, and the base and tip of the nose do not move drastically. Each landmark is located automatically on the face surface by means of a Markov network following the procedure proposed by Ben Azouz et al.~\cite{azouz06}. The network learns the statistics of a property of the surface around each landmark and the structure of the connections shown in Fig. \ref{fig:landmarks}. 

\subsection{Learning}\label{sec:learning}

\begin{figure}[htp]
\centering
\includegraphics[width=0.45\textwidth,height=!]{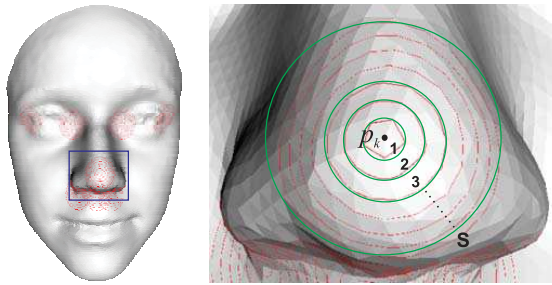}
\caption{Circles used to compute the Finger Print descriptor. Red and green circles correspond to the Geodesic and Euclidean circles, respectively.}\label{fig:fingerprint}
\end{figure}

Two important aspects have to be defined for the training of the Markov network. First, each landmark $l_i$ ($i=1,2,\ldots,L$), represented by a network node, is described using a node potential. We use a surface descriptor based on the \textit{Finger Print} (\textit{FP})~\cite{sun01}. The descriptor uses a measure related to the area of a geodesic circle centered at the point to be characterized. The descriptor at a point $p_k$ ($k=1,2,\ldots,N$, $N$ is the number of vertices in the model) is obtained by computing the distortion of the geodesic disks with respect to Euclidean disks of the same radius. More specifically, the distortion of the area $A(c)$ of the geodesic disk $c$ of radius $r$ centered at $p_k$ is computed as $d(r)=A(c)/(\pi r^2)$. We use as descriptor of $p_k$ a vector of distortions $d(r_i)$ obtained by varying the radius $r_i$ of the geodesic disk (see Fig. \ref{fig:fingerprint}). The reason we use \textit{FP} as node potential is because it is isometry-invariant. Hence, in scenarios where the surface undergoes changes that preserve isometry, $FP$ has been effective to encode the surface information of an object. \textit{FP} is used to predict landmarks on human models in varying poses~\cite{wuhrer2010b}.

Second, a link between landmarks, represented by a network edge, is described using an edge potential. Although we selected the locations of the landmarks based on the observations that nose and eye regions do not change much in the presence of expressions, some distortions along the edges of the Markov network may occur. To minimize the effects of the face movements, we compute the canonical form~\cite{elad03} of each model and define the edge potential as the relative position of landmark $l_i$ with respect to landmark $l_j$ in the canonical form space. We compute the canonical form as the embedding of the intrinsic geometry of the face surface to $\mathbb{R}^3$. To compute this embedding, we perform least-squares multi-dimensional scaling~\cite{Cox01} with geodesic distances between vertices as dissimilarities, and the geodesic distances are computed using fast marching~\cite{elad03}. We choose these standard techniques as they are efficient. These potentials ensure that the model is isometry-invariant.

The Markov network training process learns the distributions of both node and edge potentials. We assume Gaussian distributions for both the node and edge descriptors in this paper, and we learn the distributions using maximum likelihood estimation. We choose this commonly used distribution to derive an efficient algorithm that is easy to implement. While this distribution may not be satisfied in practice, we found experimentally that using this simplified assumption yields satisfactory results.

\subsection{Prediction with Belief Propagation}\label{sec:prediction}
 
The estimation of the location of landmarks on a test model is carried out by using probabilistic inference over the Markov network. In practice, we perform inference using the loopy belief propagation algorithm~\cite{yedidia03}. This algorithm requires a set of possible labels for each node. In our case, this means we need to provide a number of candidate locations for each landmark.

Wuhrer et al.~\cite{wuhrer2011a} use canonical forms to learn the average locations of the landmarks, but because of the flipping-invariant property of the canonical forms, it is necessary to compute eight different alignments and select the one that leads to the minimum distance between the scan and the deformed template. In this work, we design a method to restrict the search space based on a rough template alignment. In this way, only one fitting process has to be computed, reducing the computing cost by a factor of eight.

\subsection{Restricting the search region}\label{restricting}

There are two reasons to reduce the search space for the landmarks: to increase the efficiency of the landmark prediction and to eliminate the ambiguity caused by the facial symmetry. We treat the problem of restricting the search region for the landmarks as a 3D face pose estimation problem. In our case, the estimated pose does not have to be so accurate since the Markov network refines the position of the landmarks, but it has to be accurate enough to identify the left and right sides of the face. The proposed face pose estimation method finds four landmarks located on the nose region and extracts the information of the face symmetry planes by using a template of the landmark graph. Once the nose landmarks are labeled, the final position of the entire set of landmarks is obtained by transforming the template to the coordinate system of the test model. Fig. \ref{fig:alignment} shows the main steps of the proposed search space restriction method.

\subsection{Classification of Vertices}\label{classify}

Before explaining the rough template alignment procedure, we introduce a method to classify a vertex of a 3D model into a specific class. In our case, the classes correspond to the nodes of the Markov network and the 3D model corresponds to a 3D face model. The decision rules are derived from a clustering procedure over the Principal Components Analysis (PCA) projections of a surface feature and a pre-selection method based on the surface primitives.

\begin{figure*}[htp]
\centering
\includegraphics[width=0.98\textwidth,height=!]{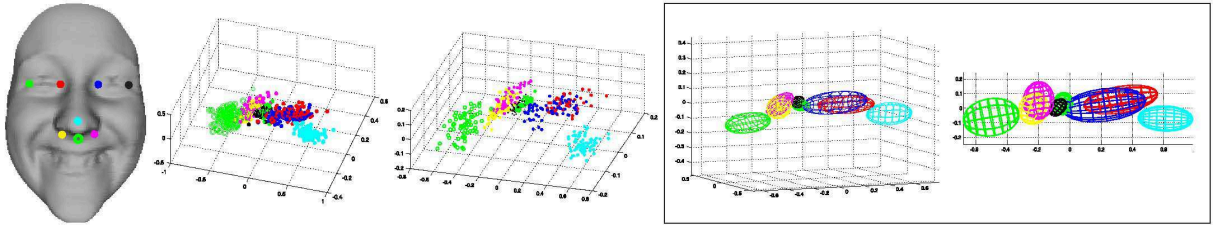}
\caption{PCA-based clustering. Left to right: Landmarks on a face model. Initial clusters formed with all the samples. Final cluster after removing the samples beyond a $1.5$ standard deviations from the cluster medoid. Minimum volume enclosing ellipsoids (3D and upper views).}\label{fig:PCA}
\end{figure*}

As the value of the $FP$ descriptor at each landmark $l_i$ was computed during the Markov network training process, we can model the distributions of the surface descriptors and use them to classify a vertex $v_k$ on the face surface into a class $i$ (each landmark corresponds to a class). PCA is a useful tool to compress a high-dimensional space into a linear low-dimensional space. When the space corresponds to a multidimensional feature space, sometimes, depending on the distinctiveness of the features, it is possible that elements of the same class form clusters in the PCA space. In our case, the $FP$ descriptor can be viewed as $S$-dimensional vector and PCA is used to reduce the dimensionality to $D$. In this work we choose $D=3$. Fig. \ref{fig:PCA} shows the results of applying PCA to the data from the subjects in neutral and performing six expressions (for information about the database, see Section \ref{DB}).

Although samples of the same class tend to form groups in the PCA space, some groups overlap due to symmetric landmarks. In order to improve the separation between classes, we define a new cluster, denoted as \textit{M-cluster}, by removing the samples which are farther than $M$ $(M\in\mathbb{R}^+)$ times the standard deviation from the cluster medoid. Medoids are representative objects of a cluster whose average dissimilarity to all the objects in the cluster is minimal~\cite{han06}. For instance, Fig. \ref{fig:PCA} shows the \textit{M-clusters} formed by setting $M=1.5$. With this value, the clusters corresponding to the landmarks nose tip and subnasal (points $7$ and $8$ in Fig.~\ref{fig:landmarks}) do not overlap any of the clusters. We will show in Sections \ref{refining} and \ref{alignment} that with a good separation between these two classes, a proper landmarks prediction can be obtained.

\begin{figure}[htp]
\centering
\includegraphics[width=0.5\textwidth,height=!]{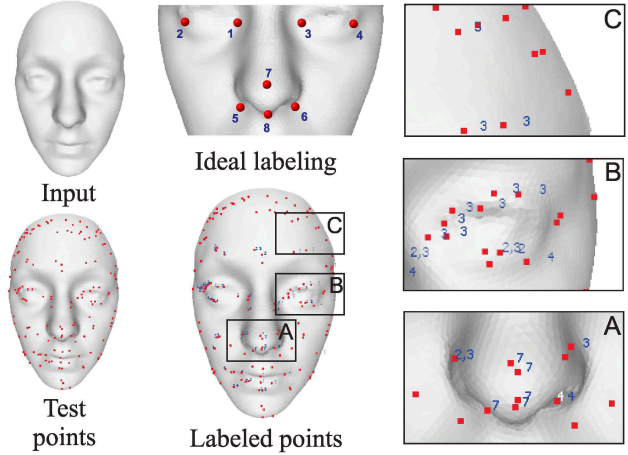}
\caption{Example of vertex labeling result. (A) Notice how the points on the nose tip region are correctly labeled. (B) Some vertices are assigned to two classes. This situation is because of the left-right symmetry of the features. (C) Points located far from the region of interest are discarded.}\label{fig:labeled}
\end{figure}

We derive a rule $E_i$ for a class $i$ based on a clustering procedure. The rule $E_i$ is defined as the minimum volume enclosing ellipsoid of a \textit{M-cluster}$_i$ (see Fig. \ref{fig:PCA}). $E_i$ is obtained from the representation of the ellipsoid in the center form as $(p_k-C_i)^T A (p_k-C_i) \leq 1$, where $C_i$ corresponds to the center of the ellipsoid corresponding to class $i$ and $A$ is the $3\times3$ matrix of the ellipse equation. When a new point $p_k$ becomes available, each $E_i$ is evaluated in order to see if the point satisfies the equation. As some \textit{M-clusters} are overlapping, it is possible that more than one label be assigned to the same $p_k$. Similarly, it is possible that $p_k$ is not assigned to any class because the point lies in a region that is not of interest. Fig. \ref{fig:labeled} shows an example of the vertex classification results obtained using the proposed method.

\begin{figure}[htp]
\centering
\includegraphics[width=0.9\textwidth]{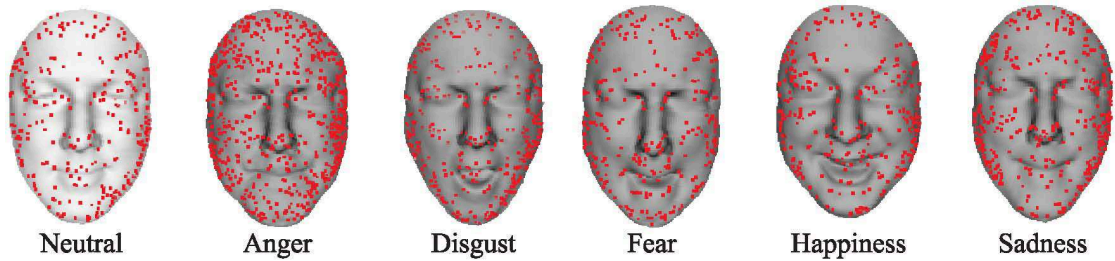}
\caption{\textit{Umbilics} of different 3D facial models of the same subject performing different expressions. Notice how the \textit{umbilics} are distributed all over the surface, and in most of the cases umbilics are present at the locations of salient facial features.}\label{fig:umbilics}
\end{figure}

It is not efficient to compute the descriptor value and its projection to PCA space for all the vertices of the mesh. To reduce the search space, we compute samples on the surface using a curvature-based descriptor. More precisely, we use as samples all surface \textit{umbilics}~\cite{cazals04b}, which are the points on the surface where the principal curvatures are identical (that is, $k_1 = k_2$). We choose this sampling approach because it can be observed experimentally that most landmark positions are located close to a umbilic, as shown in Fig. \ref{fig:umbilics}.

\subsection{Refining the Nose Landmarks}\label{refining}

\begin{figure*}[htp]
\centering
\includegraphics[width=0.80\textwidth,height=!]{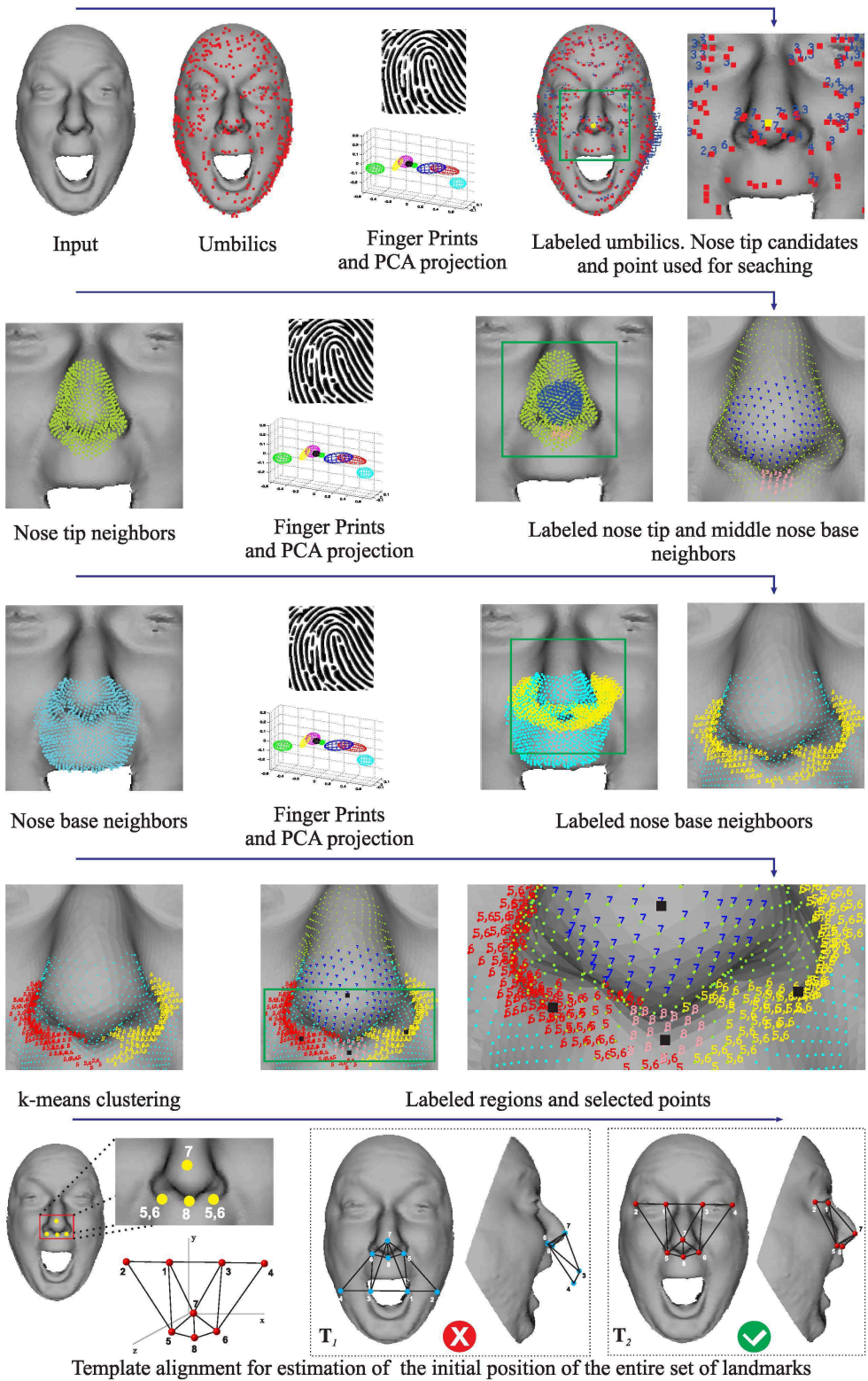}
\caption{Framework of the proposed initial alignment method.}\label{fig:alignment}
\end{figure*}

In this section we describe the procedure to select four points on the nose area, which are used as initial guess of the landmarks: right subalare, left subalare, nose tip, and subnasal, which are labeled as $5$, $6$, $7$ and $8$, respectively (see Fig.~\ref{fig:landmarks}). Following the classification procedure described in Section~\ref{classify}, for each umbilic of the input scan $F$, the $FP$ descriptor is computed, projected into PCA space, and labeled (in the following we refer to this procedure as \textit{FPPCA}). The result is a set of candidates for each landmark class (see first row of Fig.~\ref{fig:alignment}). As the umbilics of $F$ are located close to the position of the landmarks of interest, before selecting one of the labeled umbilics as landmark, it is worth to inspect their neighborhoods by looking for candidates that are closer to the landmark of interest than the initial candidate. The search starts in the nose tip region. As starting point, we select the vertex $v$ of $F$, which is the umbilic that after the \textit{FPPCA} procedure is the closest point to the medoid of the cluster corresponding to the nose tip. The new search space corresponds to the set of vertices $v_k$ within the geodesic circle of radius $r$ centered at $v$. For each vertex $v_k$, the \textit{FPPCA} procedure is applied. Since only the vertices labeled as $7$ and $8$ are considered, two new sets of candidates are obtained. This procedure is depicted in the second row of Fig.~\ref{fig:alignment}.

The search for the points labeled $5$ and $6$ is performed in a similar way as for the points labeled $7$ and $8$. In this case, the starting point corresponds to the vertex that has been labeled as $8$ in the previous step and that after the \textit{FPPCA} procedure is the closest point to the medoid of the cluster corresponding to the subnasal. In this case, only the points labeled $5$ or $6$ are considered. As mentioned before, due to the overlapping \textit{M-clusters}, most of the labeled points are assigned to two classes and the non-relevant points are discarded (see third row of Fig.~\ref{fig:alignment}). Taking into account that the labeled vertices are distributed over both sides of the nose, in order to split up this set of vertices into two sets, we perform a $k-means$ clustering over this set of vertices. The two new sets of vertices still have the same label but each of them define the neighborhood for searching either the initial guess of the point $5$ or the point $6$ (see fourth row of Fig.~\ref{fig:alignment}).

The next step is to select the four points on the nose region that will be used as initial set of landmarks. The points labeled $7$ and $8$ correspond to the points that minimize the distance to the medoid of its respective cluster after the \textit{FPPCA} procedure. The two points labeled $5$ and $6$ correspond to the vertices of $F$ that are the closest points to the centroid of each cluster resulting from the $k-means$ clustering. The proper labels of the points labeled $5$ and $6$ are derived from the procedure described in the next Section.

\subsection{Aligning Landmark Graph to Scan}\label{alignment}

So far, four points on the nose region have been selected and labeled. Due to the face symmetry, two of the points have the same labels. To solve this problem, a template $P_a$ of the upper part of the face with the same structure as the landmark graph (see Fig. \ref{fig:landmarks}) is roughly aligned to the input scan $F$. This helps also to estimate the initial guess of the remaining landmarks: right inner eye corner, right outer eye corner, left inner eye corner, and left outer eye corner, which are labeled as $1$, $2$, $3$ and $4$, respectively.

We compute a rigid alignment $\bf{T}$ that best aligns the point set $v_a$ from $P_a$ with the point set $v_b$ from $F$. The point set $v_a$ corresponds to the points labeled $5$ to $8$ of $P_a$, and $v_b$ corresponds to the four points on the nose region of $F$. As the labels $5$ and $6$ of the points in $v_b$ are unknown, there are two possible configurations for the alignment. As a result two linear transformations ${\bf{T}}_1$ and ${\bf{T}}_2$ are obtained. In order to select the transformation that produces a valid result, the transformed point sets $P_1 = {{\bf{T}}_1} P_a$, and $P_2 = {{\bf{T}}_2} P_a$, are computed. One of the transformations produces a vertical ``flip'' of the template, resulting in a wrong estimation of the coordinates of the points in the eye region. Therefore, the points set $P_i$ that minimizes the distance $D_F$ to $F$ ($D_F$ corresponds to the summation of the Euclidean distances between each point of $P_i$ and its closest point in $F$) indicates which transformation is correct. This procedure is depicted in the fifth row of Fig.~\ref{fig:alignment}.

The locations of the transformed template vertices are used to define the search space region on which statistical inference is performed. We then perform statistical inference on these search space regions using belief propagation to predict the landmarks, as discussed in Section ~\ref{sec:prediction}.


\section{Registration}
\label{sec:registration}

\begin{figure*}[htp]
\centering
\includegraphics[width=\textwidth,height=!]{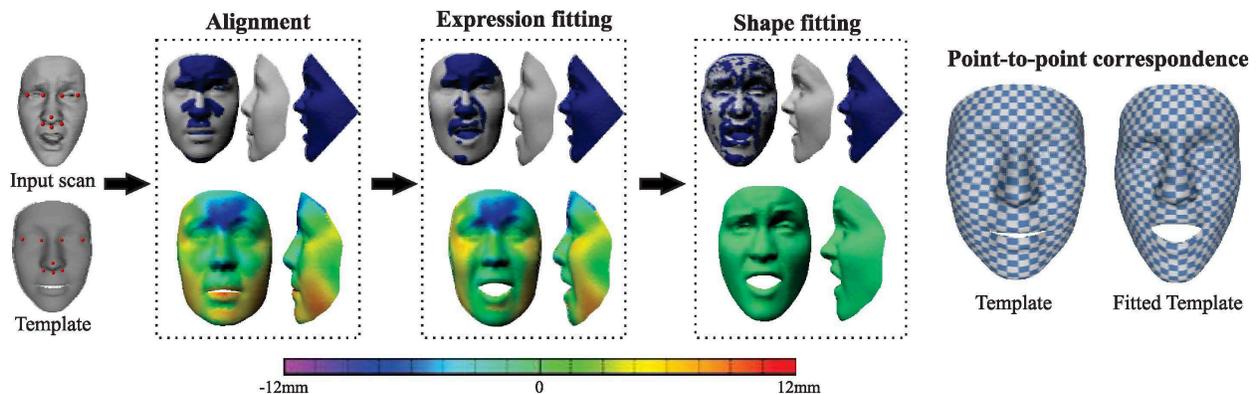}
\caption{Registration procedure. First, the template and the scan are aligned using the predicted landmarks. Second, the expression is fitted using a blendshape model. Finally, an energy-based surface fitting method is used to fit the shape. At the end, the overlap between the scan and the template is maximized and a point-to-point correspondence for the face shapes in different expressions is obtained.}\label{fig:registration}
\end{figure*}

In this section, we describe how a template is fitted to a 3D scan of the face. The input scan corresponds to a face of a subject performing a facial expression. Fitting a template to this scan is challenging because the facial geometry has large variations due to different face shapes and facial muscle movements. We propose a registration method, where the expression and the shape are fitted separately in order to handle the complexity of the problem. Fig. \ref{fig:registration} shows an overview of the proposed method.

We address the facial expression fitting problem as a facial rigging problem. In facial rigging, a facial expression is produced by changing a set of parameters associated with the different regions of the face modeled using blendshapes. Conceptually, to generate a facial shape from a 3D rest pose face template, we just move a set of vertices to a new location, e.g., lift an eyebrow or open the mouth (see Fig. \ref{fig:blendshapes}). In this sense and similar to the approach proposed by Li et al.~\cite{Li10}, we model a facial expression as a linear combination of facial blendshapes (denoted by $A_i$), which are expressed as vectors of displacements from the rest pose (denoted by $A_0$).

\subsection{Affine Alignment}\label{sec:affine}

To solve the fitting problem, the template $A_0$ in neutral pose is aligned to a scan $F$ as follows. Both $A_0$ and $F$ contain a set of landmarks denoted by ${\bar{l_i}}$ and $l_i$, respectively. The landmarks $l_i$ were predicted using the method described in Section~\ref{sec:landmark}. The alignment is carried out by finding a $3\times4$ transformation matrix ${{\bf{T}}_A}$ that minimizes the energy

\[{E_{lnd}} = \sum\limits_{i = 1}^L {\left( {{{\bf{T}}_A}{{\bar l}_i} - {l_i}} \right)}^2,\]

\noindent{with respect to the 12 parameters in ${{\bf{T}}_A}$ using a quasi-Newton approach starting from ${{\bf{T}}_A}$ as identity matrix}.

\subsection{Expression Fitting}\label{sec:expression}

We now outline how to fit the expression of the blendshape model to $F$. The aim of this step is to model expression variations. An expression can be generated as

\begin{equation}\label{eq:blendshape}
\centering
P(\alpha _i) = {A_0} + \sum\limits_{i = 1}^j {{\alpha _i}{A_i}},
\end{equation}

\begin{figure}[htp]
\centering
\includegraphics[width=0.7\textwidth,height=!]{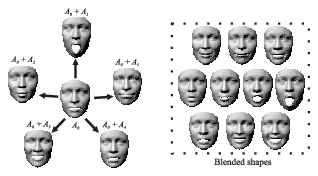}
\caption{Left: template rest pose $A_0$ and a set of blendshapes $A_i$. Right: examples of models generated as linear combinations of blendshapes.}\label{fig:blendshapes}
\end{figure}

\noindent{where $A_0$ corresponds to the rest pose, $A_i,i>0$ correspond to the blendshape displacements, and $\alpha_i \left( {0 \le {\alpha _i} \le 1} \right)$ are the blending weights of expression $P(\alpha _i)$. For each blendshape $A_i$, Fig.~\ref{fig:blendshapes} shows the corresponding expressions. The 3D models used in both the creation of $A_0$ and the generation of $A_i$ were obtained using a commercial software. Notice that mostly mouth displacements are considered. As the expressions are generated as a linear combination of displacements, to avoid exaggerated undesired expressions, it is important that no two blendshapes add the same kind of displacement. By using a blendshape model, the facial expression fitting problem is transformed into an optimization problem, where the value of each $\alpha_i$ has to be estimated.}

\begin{figure}[htp]
\centering
\includegraphics[width=0.2\textwidth]{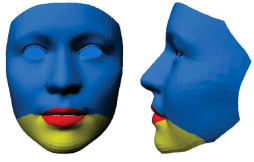}
\caption{Regions used in the expression fitting procedure.}\label{fig:region}
\end{figure}

Recall that $A_0$ and $F$ are affinely aligned. We find the $\alpha_i$ that best match the expression of $F$ by dividing $P(\alpha _i)$ into three regions: upper face, chin and mouth (as shown in Fig. \ref{fig:region}). The division is motivated by the fact that the chin and lip regions vary drastically from one expression to another (mostly in terms of displacements). Thus it is desirable to inspect the quality of the fitting in each of these regions separately. Face regions like eyebrows and cheeks also change their shapes to produce the expressions but we expect that these changes can be captured during the shape fitting step.

To fit the expression, we use the energy

\begin{equation}\label{eq:eexpression}
\centering
{E_{expr}} = \sum_r \omega_r \left\langle (NN(p_r(\alpha_i)) - p_r(\alpha_i)), \vec{n}(NN(p_r(\alpha_i)))\right\rangle^2
\end{equation}

\noindent{where $p_r(\alpha _i)$ are the vertices of $P(\alpha _i)$, $NN(p_r(\alpha_i))$ indicates the nearest neighbor of $p_r(\alpha_i)$ on $F$, $\vec{n}(NN(p_r(\alpha_i)))$ is the unit outer normal of $NN(p_r(\alpha_i))$, $\left\langle ., .\right\rangle$ denotes the dot product of two vectors, and $\omega_r$ is a weight associated with $p_r(\alpha_i)$. The energy pulls each vertex of the template to the nearest point on the tangent plane of its nearest neighbor on $F$. The weight $\omega_r$ is used for two purposes: to give different weight to the mouth and chin regions of the model, and to make the method more robust to both the presence of outliers and mis-oriented surfaces. To achieve the first goal, we set $\omega_r$ to either $\omega_{upper}, \omega_{chin},$ or $\omega_{mouth}$, depending on the region containing $p_r(\alpha _i)$. To achieve the second goal, we only consider the nearest neighbor if the angle between the outer normal vectors of $p_i(\alpha _i)$ and $NN(p_r(\alpha_i))$ is small. Specifically, we set $\omega_{upper}$ to zero if the angle is larger than $\varphi$. To force the fit to be exact, we set $\omega_{chin}$ and $\omega_{mouth}$ to zero if the angle is larger than $\varphi/2$. The expression is fitted by minimizing Eq. \ref{eq:eexpression} with respect to the blending weights $\alpha_i$. In our experiments we set $\varphi$ to 80 degrees.}

The minimization of $E_{expr}$ is carried out in two stages. In the first stage, we inspect if some movement occurs in the chin. Once we know the position of the chin, to refine the match with the expression of the input model, we need to inspect the positions of the lips. Based on this, the expression fitting procedure proceeds as follows: First, the weight $\omega_{mouth}$ is set to zero, thus the minimization is only guided by vertices that are not in the mouth region. In this step $\omega_{upper}$ is set to one and $\omega_{chin}$ is defined as $1 - \left( {{V_{chin}^{valid}}/V_{chin}} \right)$, where $V_{chin}$ is the number of vertices in the chin region and $V_{chin}^{valid}$ is the number of valid nearest neighbors in this region. The second step begins when at least $80\%$ of the vertices in the chin region have valid nearest neighbors. At this time, $\omega_{mouth}$ is set to $1 - \left( {{V_{mouth}^{valid}}/V_{mouth}} \right)$, where $V_{mouth}$ is the number of vertices in the mouth region and $V_{mouth}^{valid}$ is the number of valid nearest neighbors in this region. The minimization process ends when at least $60\%$ of the vertices in the mouth region have valid nearest neighbors. This weight variation scheme ensures that the chin and mouth regions of $P(\alpha _i)$ match the expression of $F$. The threshold values for $\omega_{chin}$ and $\omega_{mouth}$ were chosen based on experimental observations.

\subsection{Shape Fitting}\label{sec:shape}

As most of the changes in terms of movement, especially in the chin and mouth regions, were captured in the expression fitting stage, the next step consists of adapting the shape of $P(\alpha_i)$ to the shape of $F$. For ease of notation, we use $P=P(\alpha_i)$ in the following.

The shape fitting is, again, treated as an optimization problem similar to the method proposed by Allen et al.~\cite{Allen03} and extended by Li et al.~\cite{li09}. The goal is to find a set of $3\times4$ transformation matrices ${{\bf{T}}_i}$ for each vertex $p_i$ of $P$ such that $p_i$ is moved to the new location $\tilde{p}_i = {{{\bf{T}}_i}{p_i}}$ to fit the shape of $F$. The transformed version of $P$ is denoted ${\tilde P}$. The transformation matrices ${{\bf{T}}_i}$ are obtained by minimizing an energy function, which is a weighted sum of three energy terms. 

The first term is the data term 

\[{E_{data}} = \sum\limits_{i} \omega_i \left\langle \left( NN \left( \tilde{p}_i \right) - \tilde{p}_i \right), \vec{n} \left( NN \left( \tilde{p}_i \right) \right) \right\rangle^2,\]

\noindent{where ${NN\left( \tilde{p}_i \right)}$ indicates the nearest neighbor of $\tilde{p}_i$ on $F$, and $\vec{n} \left( NN \left( \tilde{p}_i \right) \right)$ is the normalized outer normal of ${NN\left( \tilde{p}_i \right)}$. The weight $\omega_i$ is set to one if the angle between the outer normal vectors of ${\tilde p_i}$ and its nearest neighbor is at most 80 degrees, and to zero otherwise. The data term ensures that the template is deformed to resemble the input scan.}

The second energy is a smoothness term that encourages smooth transformations between neighboring vertices of the mesh. We define it as

\[{E_{smooth}} = \sum\limits_{(i,j) \in {E^{\left( {\tilde P} \right)}}} {{{\left( {{{\bf{T}}_i} - {{\bf{T}}_j}} \right)}^2}},\]

where ${{E^{\left( {\tilde P} \right)}}}$ is the  set of edges of ${\tilde P}$. This term prevents adjacent parts of $P$ from being mapped to disparate parts of $F$, and also encourages similarly-shaped features to be mapped to each other~\cite{Allen03}.

The final energy term encourages the transformation matrices to be rigid. The rigid energy $E_{rigid}$, which measures the deviation of the column vectors of ${{{\bf{T}}_i}}$ from orthogonality and unit length, is defined as

\[\begin{array}{l}
 {E_{rigid}} = \sum\limits_{i = 1}^r {\left( {{{\left( {{{\left( {{\bf{a}}_1^i} \right)}^{\rm{T}}}{\bf{a}}_2^i} \right)}^2} + {{\left( {{{\left( {{\bf{a}}_1^i} \right)}^{\rm{T}}}{\bf{a}}_3^i} \right)}^2} + {{\left( {{{\left( {{\bf{a}}_2^i} \right)}^{\rm{T}}}{\bf{a}}_3^i} \right)}^2}} \right.}  +  \\ 
 \,\,\,\,\,\,\,\,\,\,\,\,\,\,\,\,\,\,\,\,\,\,\,\,\,\,\,\,\,\left. {\,\,\,\,\,{{\left( {1 - {{\left( {{\bf{a}}_1^i} \right)}^{\rm{T}}}{\bf{a}}_1^i} \right)}^2} + {{\left( {1 - {{\left( {{\bf{a}}_2^i} \right)}^{\rm{T}}}{\bf{a}}_2^i} \right)}^2} + {{\left( {1 - {{\left( {{\bf{a}}_3^i} \right)}^{\rm{T}}}{\bf{a}}_3^i} \right)}^2}} \right), \\ 
 \end{array}\]
 
\noindent{where ${{\bf{a}}_1^i}$, ${{\bf{a}}_2^i}$, ${{\bf{a}}_3^i}$ are the first three columns vectors of ${{{\bf{T}}_i}}$.}

The energy terms described above are combined in the weighted sum

\begin{equation}\label{eq:eshape}
\centering
{E_{shape}} = \omega_{data} {E_{data}} + \omega_{smooth} {E_{smooth}} + \omega_{rigid} {E_{rigid}}.
\end{equation}

The shape is fitted by minimizing $E_{shape}$ with respect to the parameters ${{{\bf{T}}_i}}$. We start by encouraging smooth and rigid transformations by setting $\omega_{data}=1$, $\omega_{smooth}^0=20000$, and $\omega_{rigid}^0=10$. Similar to Li et al.~\cite{li09}, whenever the energy change is negligible, we relax the weights as  $\omega_{smooth}^t=0.5\omega_{smooth}^{t-1}$ and $\omega_{rigid}^t=0.5\omega_{rigid}^{t-1}$ to give more weight to the data term. This allows the template to deform towards the scan. The algorithm iterates until the relative change in energy $(E_{shape}^{i-1} -E_{shape}^i) /E_{shape}^{i-1}$, where $i$ is the iteration number, is less than $0.0001$. For each set of weights, we use a quasi-Newton approach~\cite{liu89} to solve the optimization problem, and we perform at most $1000$ iterations. 

As our template only captures the generic shape and deformation model of a face and can be freely deformed during shape fitting, points on the boundary of the input model are ignored as nearest neighbors to prevent including matches from the hair or ears of the scan.

\section{Experiments and results}
\label{sec:experiment}

\subsection{Database}\label{DB}

\begin{figure*}[htp]
\centering
\includegraphics[width=\textwidth,height=!]{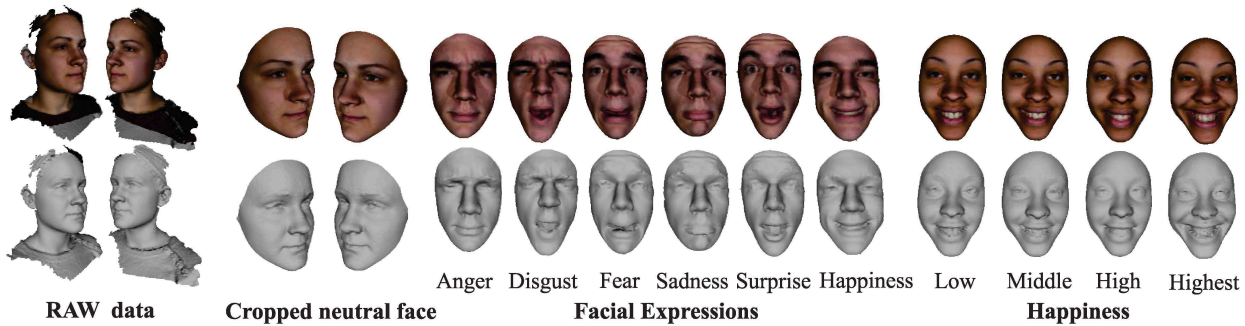}
\caption{Characteristics of the BU-3DFE database.}\label{fig:BU3DFEsamples}
\end{figure*}

We use the BU-3DFE~\cite{yin06} database for our experiments. The database consists of 3D face models from 100 subjects (56 Females and 44 Males) in neutral pose and with the following facial expressions: \textit{surprise}, \textit{happiness}, \textit{disgust}, \textit{sadness}, \textit{anger} and \textit{fear}. There are four scans of each facial expression, corresponding to different levels of intensity from \textit{low} to \textit{highest}. As a file containing the raw data of each scan is also available, there are a total of 50 files per subject, 25 raw scans and 25 corresponding to the cropped faces. Fig. \ref{fig:BU3DFEsamples} shows snapshots of different scans from the BU-3DFE database. In this work, we use a subset of 700 3D models corresponding to the cropped faces of the subjects performing the expressions in the highest level.

\subsection{Landmark prediction accuracy}\label{sec:accuracy}

We use two different subsets of models of 50 subjects (25 females and 25 males) to train the landmark prediction model. First, we use a subset $T_n$ consisting of 50 models of subjects in neutral pose as training set. Second, we use a subset $T_e$ consisting of 350 models of subjects in neutral pose and performing six different facial expressions as training set. As $T_n$ covers the shape variability and $T_e$ covers both shape and expressions variability, we are able to evaluate the importance of the variabilities considered in the training sets. The accuracy of the landmark prediction algorithm is evaluated over the remaining 50 subjects of the database (31 females and 19 males). The test database corresponds to 350 models of subjects in both neutral pose and when performing six different facial expressions.

To evaluate the accuracy of the landmark prediction algorithm, we compute the error of the Euclidean distance between a manually located landmark $l_i$ and its corresponding estimation ${\hat l_i}$. We compute the mean, the standard deviation and the maximum of the error. We also compute the detection rates by counting the percentage of test models where the landmark ${\hat l_i}$ was predicted with an error below $10mm$ ($T10$), $20mm$ ($T20$), and $30mm$ ($T30$). Tables \ref{tab:errorTn} and \ref{tab:errorTe} show the results of the evaluation for the test with $T_n$ and $T_e$ as training databases, respectively. 

\begin{table}
\centering
\begin{tabular}{c c c c c c}
\toprule
\textbf{Landmark} & \textbf{Mean $\pm$ Std} & \textbf{Max.} & \textbf{$T10$} & \textbf{$T20$} & \textbf{$T30$} \\
         &   [mm]  & [mm] &   [\%]  & [\%] &   [\%]  \\
\toprule
Right inner eye corner	&	10.35	$\pm$	6.13	&	33.93	&	53.71	&	87.14	&	92.57	\\
\midrule
Right outer eye corner	&	11.79	$\pm$	7.77	&	34.73	&	27.71	&	85.71	&	93.43	\\
\midrule
Left inner eye corner	&	11.63	$\pm$	6.82	&	34.16	&	44.57	&	86.57	&	94.00	\\
\midrule
Left outer eye corner	&	12.57	$\pm$	7.23	&	34.29	&	31.43	&	89.14	&	95.71	\\
\midrule
Right subalare	&	9.96	$\pm$	6.59	&	33.49	&	66.00	&	86.86	&	98.00	\\
\midrule
Left subalare	&	10.93	$\pm$	6.87	&	34.15	&	55.14	&	87.43	&	94.29	\\
\midrule
Nose tip	&	7.42	$\pm$	5.64	&	32.03	&	82.57	&	92.00	&	96.86	\\
\midrule
Subnasal	&	7.12	$\pm$	5.87	&	33.75	&	84.57	&	87.43	&	95.43	\\
\bottomrule
\end{tabular}
\caption{Error of landmark prediction with training set $T_n$. $T10$, $T20$, and $T30$ correspond to the detection rates with a tolerance of $10mm$, $20mm$ and $30mm$, respectively.}
\label{tab:errorTn}
\end{table}

The best landmark prediction results were obtained when $T_e$ is used for training. In both experiments, the landmarks located in the nose region are better predicted than the ones located in the eye region. The tip of the nose is predicted with the lowest error and the outer corners of the eyes are predicted with the highest error. One of the reasons that the outer corners of the eyes are not predicted as well as the other landmarks is that the initial position is found based on the alignment of the landmark template (see Fig. \ref{fig:alignment}). This adds an estimation error that is reflected in the  values of the standard deviation. The values of the detection rates show the improvement in accuracy of the landmark prediction when $T_e$ is used as training set. This indicates that for the configuration of the landmark prediction model used in this work, the variations due to both shape and expression have to be considered.

\begin{table}
\centering
\begin{tabular}{c c c c c c}
\toprule
\textbf{Landmark} & \textbf{Mean $\pm$ Std} & \textbf{Max.} & \textbf{$T10$} & \textbf{$T20$} & \textbf{$T30$} \\
         &   [mm]  & [mm] &   [\%]  & [\%] &   [\%]  \\
\toprule
Right inner eye corner	&	6.14	$\pm$	4.54	&	34.39	&	80.86	&	95.14	&	97.43	\\
\midrule
Right outer eye corner	&	8.49	$\pm$	6.12	&	34.54	&	62.29	&	95.14	&	97.71	\\
\midrule
Left inner eye corner	&	6.75	$\pm$	4.21	&	33.75	&	84.00	&	96.57	&	98.29	\\
\midrule
Left outer eye corner	&	9.63	$\pm$	5.82	&	34.63	&	63.14	&	93.43	&	98.86	\\
\midrule
Right subalare	&	7.17	$\pm$	3.3	&	32.23	&	85.43	&	95.14	&	97.43	\\
\midrule
Left subalare	&	6.47	$\pm$	3.07	&	32.3	&	89.71	&	96.86	&	97.43	\\
\midrule
Nose tip	&	5.87	$\pm$	2.7	&	29.91	&	93.71	&	97.43	&	100	\\
\midrule
Subnasal	&	5.57	$\pm$	2.03	&	30.26	&	95.43	&	98.29	&	99.71	\\
\bottomrule
\end{tabular}
\caption{Error of landmark prediction with training set $T_e$. $T10$, $T20$, and $T30$ correspond to the detection rates with a tolerance of $10mm$, $20mm$ and $30mm$, respectively.}
\label{tab:errorTe}
\end{table}

We compared our results of landmark prediction with two approaches where the BU-3DFE database is also used for testing. Segundo et al.~\cite{segundo10} used 2500 range images obtained from the raw data, and Nair and Cavallaro~\cite{nair09} used 2350 of the 2500 3D cropped face models available. Table \ref{tab:errorC} shows the mean of the error of the landmark prediction. For all the landmarks, our approach  outperforms the approach of Nair and Cavallaro~\cite{nair09}. Compared to Segundo et al.~\cite{segundo10}, for all the landmarks but the nose tip the mean error is similar. Recall however that Segundo et al.~\cite{segundo10} use a more challenging dataset for testing.

\begin{table}
\centering
\begin{tabular}{c c c c}
\toprule
\textbf{Landmark} &  \textbf{\cite{segundo10}} & \textbf{\cite{nair09}} & \textbf{Our Method} \\
         & [mm] & [mm] & [mm]\\
\toprule
Right inner eye corner	&	6.33	&	20.46	&	6.14\\
\midrule
Right outer eye corner	&	N.A.	&	12.11	&	8.49\\
\midrule
Left inner eye corner	&	6.33	&	19.38	&	6.75\\
\midrule
Left outer eye corner	&	N.A.	&	11.89	&	9.63\\
\midrule
Right subalare	&	6.49	&	N.A.	&	7.17\\
\midrule
Left subalare	&	6.66	&	N.A.	&	6.47\\
\midrule
Nose tip	&	1.87	&	8.83	&	5.87\\
\midrule
Subnasal	&	N.A.	&	N.A.	&	5.57\\
\bottomrule
\end{tabular}
\caption{Comparison of mean errors of our method and the approaches of Segundo et al.~\cite{segundo10} and Nair and Cavallaro~\cite{nair09}.}
\label{tab:errorC}
\end{table}

\begin{figure*}[htp]
\centering
\includegraphics[width=0.95\textwidth,height=!]{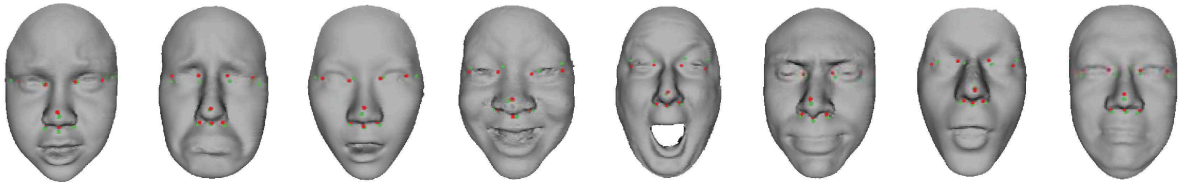}
\caption{Examples of the landmark prediction results. Red and green spheres correspond to the manually placed and predicted landmarks, respectively. First row: female subjects; Second row: male subjects.}\label{fig:exalandmarks}
\end{figure*}

Although the obtained landmark prediction error appears to be high, it is still possible to obtain a proper point-to-point correspondence since the landmarks only provide a guidance for the deformation algorithm. Fig. \ref{fig:exalandmarks} shows some examples of the landmark prediction results over models of subjects with different facial shapes and performing different expressions. 

In the following, we use $T_e$ as training dataset. Furthermore, we only consider the models where all landmarks are predicted within  $30mm$ of the ground truth (332 of the 350 models).

\subsection{Registration}\label{subsec:registration}

We tested our dense point-to-point correspondence algorithm on $332$ models. To evaluate the accuracy of the registration we computed the error in the location of manually placed landmark points present in the BU-3DFE database that are not considered for the alignment. We compute the error as the Euclidean distance between a manually placed point and its corresponding location after registration. The set of points considered for the evaluation includes $20$ points on the eyebrows ($10$ left, $10$ right), $12$ points on the eye contours ($6$ left, $6$ right), $12$ points in the nose region, and $12$ points on the outer contour of the lips. Table \ref{tab:errorTr} shows for the set of points used for evaluation, the mean, the standard deviation, and the maximum of the error, as well as the detection rates. In this case, we compute the mean and standard deviation over all points in a region and over all $332$ models used for correspondence computation. Furthermore, we compute the detection rates by counting the percentage of test models where all the points belonging to the same region were predicted with an error below $10mm$ ($T10$), $20mm$ ($T20$), and $30mm$ ($T30$). The points on the eye contour and the nose region were found with lower mean error and variation than the points on the mouth and eyebrows regions. This situation is expected because the movements in the eyebrows and mouth are more pronounced than in the other areas.

\begin{table}
\centering
\begin{tabular}{c c c c c c c}
\toprule
\multicolumn{2}{c}{\textbf{Points}} & \textbf{Mean $\pm$ Std} & \textbf{Max.} & \textbf{$T10$} & \textbf{$T20$} & \textbf{$T30$} \\
 &                  & [mm] & [mm] & [\%]  & [\%] &   [\%]  \\
\toprule
\multirow{6}*{\includegraphics[width=2.7cm]{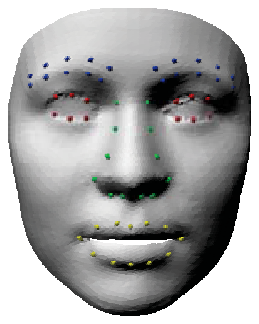}} & Left Eyeb.	& 6.28	$\pm$ 3.30	& 25.36 & 52.87 & 98.79 & 100\\
\cmidrule(r){2-7}
 & Right Eyeb.	& 6.75 $\pm$ 3.51 & 23.59 & 45.62 & 98.19 & 100 \\ 
\cmidrule(r){2-7}
 & Left Eye	& 3.25 $\pm$ 1.84 & 12.53 & 98.19 & 100	& 100 \\
\cmidrule(r){2-7}
 & Right Eye	& 3.81 $\pm$ 2.06 & 12.24 & 96.07 & 100	& 100 \\
\cmidrule(r){2-7}
 & Nose	& 3.96 $\pm$ 2.22 & 16.97 & 87.61 & 100	& 100 \\
\cmidrule(r){2-7}
 & Mouth	& 5.69 $\pm$ 4.45 & 45.36 & 52.57 & 94.26 & 98.79 \\
\bottomrule
\end{tabular}
\caption{Error at landmarks not used for registration. Left: set of points. Right: summary of errors.}
\label{tab:errorTr}
\end{table}

\begin{figure}[htp]
\centering
\includegraphics[width=0.95\textwidth,height=!]{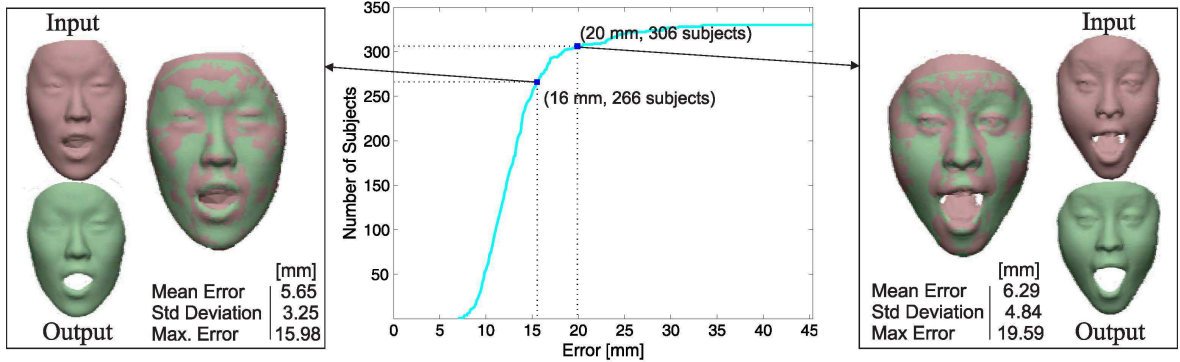}
\caption{Center: cumulative distribution of the number of models where the error at all the landmarks not used for registration is below a threshold. Left and right: example of registration results.}\label{fig:maxerrordist}
\end{figure}

Next, we discuss the quality of the results after the final shape fitting step. Fig. \ref{fig:maxerrordist} shows the cumulative distribution of the number of models where the error at all the landmarks not used for registration is below a threshold. 

We also compute the Modified Hausdorff Distance (\textit{MHD}), which is a metric for shape comparison that measures the degree of mismatch between two points sets. Therefore, it is useful to demonstrate the quality of a registration algorithm~\cite{passalis11}. The \textit{MHD} is defined as~\cite{gao03}:

\[MHD(P,F) = \frac{1}{N_p}\sum\limits_{i = 1}^{N_p} {\mathop {\min }\limits_{{{\rm{f}}_j} {\in {F}}} } \left| {{p_i} - {{\mathop{\rm f}\nolimits} _j}} \right|,\]

\noindent{where $\left| {{p_i} - {{\mathop{\rm f}\nolimits} _j}} \right|$ is the Euclidean distance between vertices of $P$ and $F$, and $N_p$ is the number of vertices of $P$. The values of the average, standard deviation and maximum of the $MHD$ for the $332$ tested models were $1.42mm$, $0.56mm$ and $3.66mm$, respectively. In addition, Fig.~\ref{fig:distances} shows the false color visualization and histograms of the mean magnitude and standard deviation of the distance between the surfaces $F$ and $P$ computed over all 332 models. For every point $P$, we consider the point-to-plane distance to the tangent plane of its nearest neighbor on $F$ (this is the distance measure used in $E_{data}$). As most of the values of the distances are concentrated between $0$ and $1mm$, in order to improve the visualization, the color map was clamped to this range. Notice the slightly high variation in the lower lip and chin area, which are the regions where the surface is deformed most due to the facial expressions. 

\begin{figure}[htp]
\centering
\includegraphics[width=\textwidth,height=!]{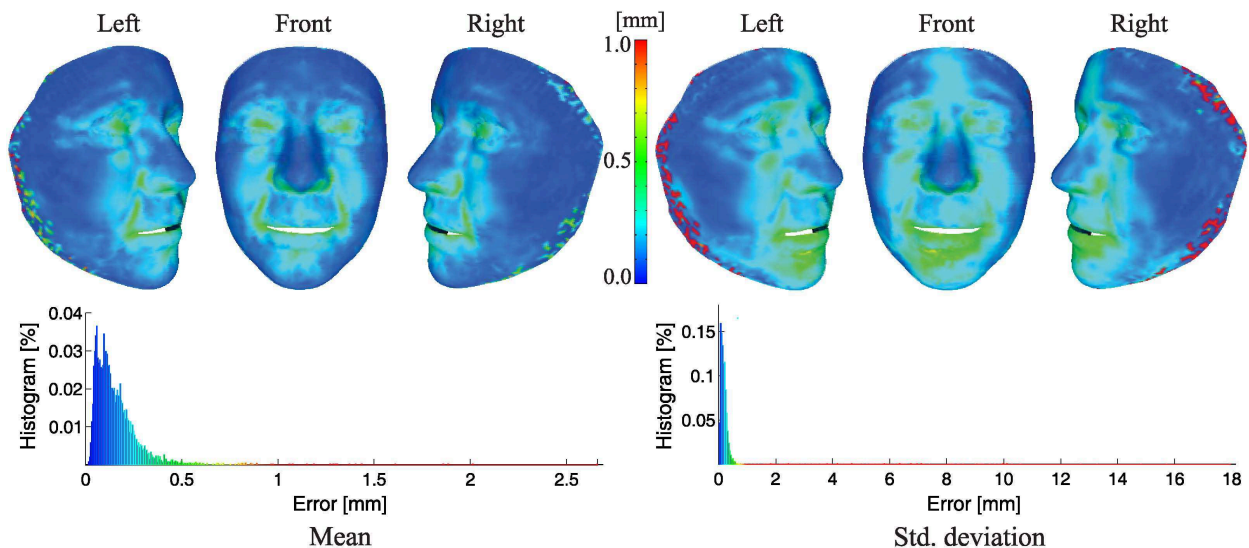}
\caption{Distance between the surfaces of $P$ and $F$. False color visualization (top) and histograms (bottom) of the magnitude of the mean and standard deviation of the distance.}\label{fig:distances}
\end{figure}

Next, we show some examples that summarize the results of the expression and shape matching stages of the registration process.
The third column of Fig. \ref{fig:exaregistration} shows examples of the expression fitting results for six different kinds of facial expression. In all cases, the expression of the mouth region of the input model is properly matched after linear blending. The fourth column of Fig. \ref{fig:exaregistration} shows examples of the shape fitting results. The models are color-coded with respect to the signed distance from the input scan. Note that most points on the models are within $2mm$ of the scan. Furthermore, notice how the different expressions in the eyebrows are properly fitted. In order to visualize the quality of the correspondences, a texture was applied to the template model (see right of Fig. \ref{fig:exaregistration}). Results of texture transferring show that in most of the face regions, the shape of the deformed template matches the shape of the input model. 

\begin{figure*}[htp]
\centering
\includegraphics[width=0.70\textwidth,height=!]{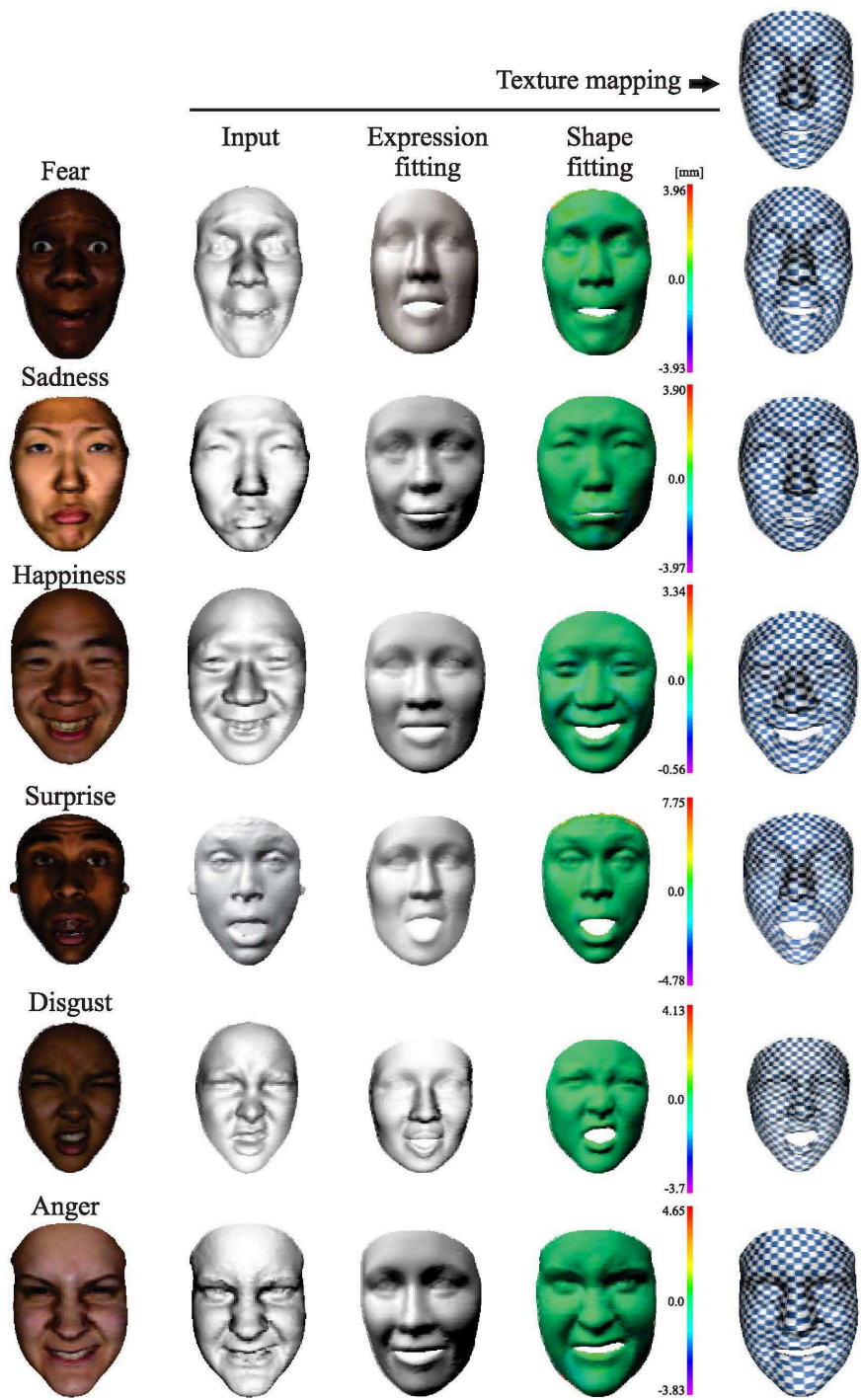}
\caption{Examples of registration results. The input, fitted expression, error mapped, and texture mapped models are provided for each example.}\label{fig:exaregistration}
\end{figure*}

We also run tests to verify if the level of the expression affects the quality of the fitting. Fig. \ref{fig:levels} depicts how the proposed method is able to correctly fit the template to different levels of expressions. For each example, the input, output and textured models are provided. Notice how both slight and pronounced movements of the eyebrows and mouth are properly matched.

\begin{figure*}[htp]
\centering
\includegraphics[width=0.6\textwidth,height=!]{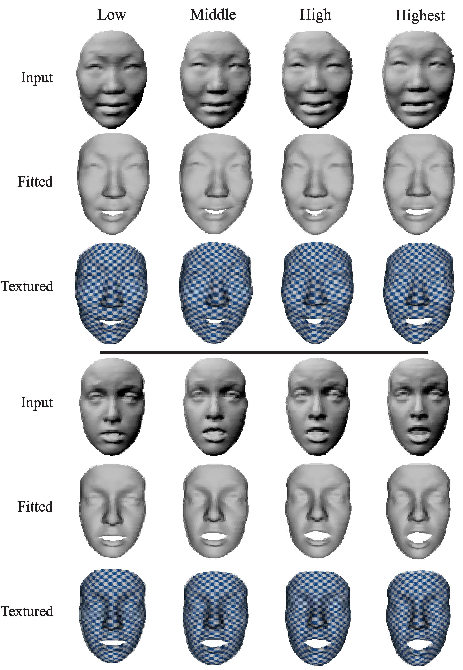}
\caption{Results of fitting to models of the same subject performing an expression in different levels. Fear (first three rows). Surprise (last three rows). For each example, first, second, and third rows are the input, output, and textured models, respectively.}\label{fig:levels}
\end{figure*}

Most of the incorrect shape fitting occurs on the inner parts of the lips. As the input scans have information in the area of the teeth, which is not considered in the template model, the algorithm converges to this region, thereby causing miscorrespondences during the shape fitting. Fig. \ref{fig:limitation} shows an example of the limitations in the shape fitting. Notice how the expression is matched correctly, but the corners of the mouth are not well located, which causes an incorrect fitting on the mouth and chin regions. 

\begin{figure}[htp]
\centering
\includegraphics[width=\textwidth,height=!]{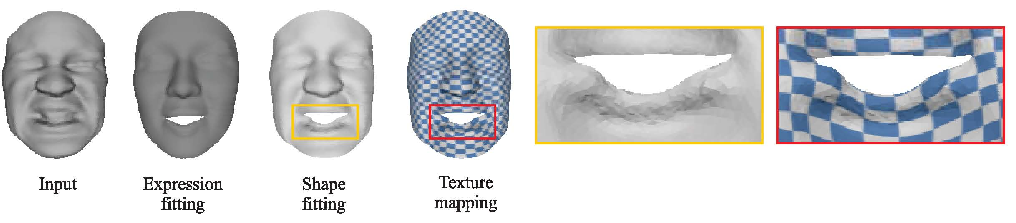}
\caption{Incorrect shape fitting. The differences in topology of the input and template meshes cause incorrect expression and shape fitting.}\label{fig:limitation}
\end{figure}

Additional tests were performed over models with occluded parts. In this case, the template was correctly fitted when the occlusion did not occur in the locations of the landmarks used for the initial alignment. Fig. \ref{fig:bosphorus} shows the result of the proposed point-to-point correspondence approach for a model of a subject where the mouth is occluded by a hand. Note that a proper registration result is obtained.

\begin{figure}[htp]
\centering
\includegraphics[width=0.4\textwidth,height=!]{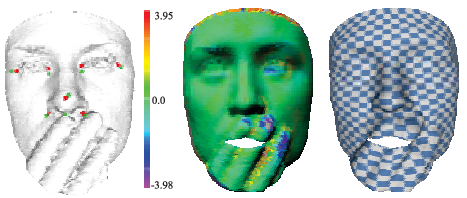}
\caption{Challenging test scenario. Mapped error models correspond to the fitting result. Test was carried out over one model of the Bosphorus database~\cite{bosphorusDB}.}\label{fig:bosphorus}
\end{figure}

Finally, we discuss the running time of our method. On a standard PC ($2.4$ GHz processor), the typical time to predict the set of landmarks for the initial alignment is about $5$ seconds for rough alignment and about $176$ seconds for the refinement of the position. The typical time for expression and shape fitting is about $6$ seconds and $28$ seconds, respectively.

\section{Conclusions}
\label{sec:conclusion}

This paper presented a fully automatic method to compute dense point-to-point correspondences between a set of human face scans with varying expressions. The proposed approach proceeds by learning local shape descriptors and spatial relationships for a set of landmark points. For a new scan, the approach first predicts the landmark points by performing statistical inference on the learned model. The approach then fits a template to the scan in two stages. The first stage fits the expression of the template to the expression of the scan using the predicted landmark points. The second stage fits the shape of the template to the shape of the scan using a non-rigid iterative closest point technique. We applied our approach to 350 models of the BU-3DFE database, and evaluated the results both qualitatively and quantitatively. We showed that for $94.9$\% of the models, the landmarks are predicted with an error below $30mm$, and that for most of the models, a consistent correspondence is found. Furthermore, we evaluated the algorithm on a challenging case of a face with occlusion.

The failure cases of the algorithm are mostly caused by noisy data in the mouth area. For future work we plan to design algorithms that can handle this challenging scenario. We will also test the algorithm on a large database of models with different types of occlusion, such as models wearing eyeglasses.

\section*{Acknowledgment}
This work was supported by the program ``\textit{Cr\'editos condonables para estudiantes de Doctorado}'' from  \textit{COLCIENCIAS}, by the program ``\textit{Convocatoria de apoyo a tesis de posgrado - Doctorados}'' from \textit{Direcci\'on de Investigaciones de Manizales}, and by the Cluster of Excellence \textit{Multimodal Computing and Interaction} within the Excellence Initiative of the German Federal Government. We thank Jonathan Boisvert, Timo Bolkart, Alan Brunton, and Pengcheng Xi for helpful discussions.


\begin{thebibliography}{10}

\bibitem{Allen03}
B.~Allen, B.~Curless, and Z.~Popovi\'{c}.
\newblock The space of human body shapes: Reconstruction and parametrisation
  from range scans.
\newblock {\em ACM Transactions on Graphics (SIGGRAPH)}, 22(3):587--594, 2003.

\bibitem{amberg08}
B.~Amberg, R.~Knothe, and T.~Vetter.
\newblock Expression invariant 3{D} face recognition with a morphable model.
\newblock In {\em IEEE International Conference on Automatic Face Gesture
  Recognition}, pages 1--6, 2008.

\bibitem{basso06}
C.~Basso, P.~Paysan, and T.~Vetter.
\newblock Registration of expressions data using a 3{D} morphable model.
\newblock In {\em International Conference on Automatic Face and Gesture
  Recognition}, pages 205--210, 2006.

\bibitem{azouz06}
Z.~Ben~Azouz and A.~Shu, C.and~Mantel.
\newblock Automatic locating of anthropometric landmarks on 3{D} human models.
\newblock In {\em International Symposium on 3{D} Data Processing,
  Visualization, and Transmission}, pages 750--757, 2006.

\bibitem{berretti11}
S.~Berretti, B.~Ben~Amor, M.~Daoudi, and A.~del Bimbo.
\newblock 3{D} facial expression recognition using {SIFT} descriptors of
  automatically detected keypoints.
\newblock {\em The Visual Computer}, 27(11):1021--1036, 2011.

\bibitem{Blanz99}
V.~Blanz and T.~Vetter.
\newblock A morphable model for the synthesis of 3{D} faces.
\newblock In {\em Conference on Computer Graphics and Interactive Techniques},
  pages 187--194, 1999.

\bibitem{Bronstein06}
A.~Bronstein, M.~Bronstein, and R.~Kimmel.
\newblock Generalized multidimensional scaling: a framework for
  isometry-invariant partial surface matching.
\newblock {\em Proceedings of the National Academy of Sciences},
  103(5):1168--1172, 2006.

\bibitem{Bronstein07}
A.~Bronstein, M.~Bronstein, and R.~Kimmel.
\newblock Expression-invariant representations of faces.
\newblock {\em IEEE Transactions on Image Processing}, 16(1):188--197, 2007.

\bibitem{Brunton11}
A.~Brunton, C.~Shu, J.~Lang, and E.~Dubois.
\newblock Wavelet model-based stereo for fast, robust face reconstruction.
\newblock In {\em Canadian Conference on Computer and Robot Vision}, pages
  347--354, 2011.

\bibitem{cazals04b}
F.~Cazals and M.~Pouget.
\newblock Smooth surfaces, umbilics, lines of curvatures, foliations, ridges
  and the medial axis: a concise overview.
\newblock Technical Report RR-5138, INRIA, 2004.

\bibitem{chang06}
K.~Chang, K.~Bowyer, and P.~Flynn.
\newblock Multiple nose region matching for 3{D} face recognition under varying
  facial expression.
\newblock {\em IEEE Transactions on Pattern Analysis and Machine Intelligence},
  28(10):1695--1700, 2006.

\bibitem{Cox01}
T.~Cox and M.~Cox.
\newblock {\em Multidimensional Scaling, Second Edition}.
\newblock Chapman \& Hall CRC, 2001.

\bibitem{Creusot2010}
C.~Creusot, N.~Pears, and J.~Austin.
\newblock {3{D}} face landmark labelling.
\newblock In {\em Proceedings ACM workshop on 3{D} object retrieval}, pages
  27--32, 2010.

\bibitem{Dryden02}
I.~Dryden and K.~Mardia.
\newblock {\em Statistical Shape Analysis}.
\newblock Wiley, 2002.

\bibitem{elad03}
A.~Elad and R.~Kimmel.
\newblock On bending invariant signatures for surfaces.
\newblock {\em IEEE Transactions on Pattern Analysis and Machine Intelligence},
  25(10):1285--1295, 2003.

\bibitem{gao03}
Y.~Gao.
\newblock Efficiently comparing face images using a modified hausdorff
  distance.
\newblock In {\em IEEE Conference on Vision, Image and Signal Processing},
  pages 346--350, 2003.

\bibitem{han06}
J.~Han and M.~Kamber.
\newblock {\em Data Mining: Concepts and Techniques, 2nd ed.}
\newblock Morgan Kaufmann Publishers, 2006.

\bibitem{Jiang05}
D.~Jiang, Y.~Hu, S.~Yan, L.~Zhang, H.~Zhang, and W.~Gao.
\newblock Efficient 3{D} reconstruction for face recognition.
\newblock {\em Pattern Recognition}, 38(6):787--798, 2005.

\bibitem{kakadiaris07}
I.~Kakadiaris, G.~Passalis, G.~Toderici, M.~Murtuza, L.~Yunliang,
  N.~Karampatziakis, and T.~Theoharis.
\newblock Three-dimensional face recognition in the presence of facial
  expressions: An annotated deformable model approach.
\newblock {\em IEEE Transactions on Pattern Analysis and Machine Intelligence},
  29(4):640--649, 2007.

\bibitem{Lu06a}
Xiaoguang L. and A.~Jain.
\newblock Deformation modeling for robust 3{D} face matching.
\newblock In {\em IEEE Computer Society Conference on Computer Vision and
  Pattern Recognition}, volume~2, pages 1377--1383, 2006.

\bibitem{lu06b}
Xiaoguang L., A.~Jain, and D.~Colbry.
\newblock Matching 2.5{D} face scans to 3d models.
\newblock {\em IEEE Transactions on Pattern Analysis and Machine Intelligence},
  28(1):31--43, 2006.

\bibitem{li09}
H.~Li, B.~Adams, L.~Guibas, and M.~Pauly.
\newblock Robust single-view geometry and motion reconstruction.
\newblock {\em ACM Transactions on Graphics (SIGGRAPH Asia)},
  28(5):175:1--175:10, 2009.

\bibitem{Li10}
H.~Li, T.~Weise, and M.~Pauly.
\newblock Example-based facial rigging.
\newblock {\em ACM Transactions on Graphics (SIGGRAPH)}, 29(4):32:1--32:6,
  2010.

\bibitem{liu89}
D.~Liu and J.~Nocedal.
\newblock On the limited memory {BFGS} method for large scale optimization.
\newblock {\em Mathematical Programming}, 45:503--528, 1989.

\bibitem{lu06}
X.~Lu and A.~Jain.
\newblock Automatic feature extraction for multiview 3{D} face recognition.
\newblock In {\em International Conference on Automatic Face and Gesture
  Recognition}, pages 585--590, 2006.

\bibitem{Mehryar10}
S.~Mehryar, K.~Martin, K.~Plataniotis, and S.~Stergiopoulos.
\newblock Automatic landmark detection for 3{D} face image processing.
\newblock In {\em IEEE Congress on Evolutionary Computation}, pages 1--7, 2010.

\bibitem{nair09}
P.~Nair and A.~Cavallaro.
\newblock 3-{D} face detection, landmark localization, and registration using a
  point distribution model.
\newblock {\em IEEE Transactions on Multimedia}, 11(4):611--623, 2009.

\bibitem{passalis11}
G.~Passalis, P.~Perakis, T.~Theoharis, and I.~Kakadiaris.
\newblock Using facial symmetry to handle pose variations in real-world 3{D}
  face recognition.
\newblock {\em IEEE Transactions on Pattern Analysis and Machine Intelligence},
  33(10):1938--1951, 2011.

\bibitem{perakis10}
P.~Perakis, G.~Passalis, T.~Theoharis, and I.~Kakadiaris.
\newblock 3{D} facial landmark detection \& face registration: A 3{D} facial
  landmark model \& 3{D} local shape descriptors approach.
\newblock Technical Report TP--2010--01, Computer Graphics Laboratory,
  University of Athens, 2010.

\bibitem{perakis09}
P.~Perakis, T.~Theoharis, G.~Passalis, and I.~Kakadiaris.
\newblock Automatic 3{D} facial region retrieval from multi-pose facial
  datasets.
\newblock In {\em Eurographics Workshop on 3{D} Object Retrieval}, pages
  37--44, 2009.

\bibitem{Romdhani02}
S.~Romdhani, V.~Blanz, and T.~Vetter.
\newblock Face identification by fitting a 3{D} morphable model using linear
  shape and texture error functions.
\newblock In {\em IEEE International Conference on Computer Vision}, pages
  3--19, 2002.

\bibitem{Romdhani03}
S.~Romdhani and T.~Vetter.
\newblock Efficient, robust and accurate fitting of a 3{D} morphable model.
\newblock In {\em IEEE International Conference on Computer Vision}, volume~1,
  pages 59--66, 2003.

\bibitem{bosphorusDB}
A.~Savran, N.~Aly\"{u}z, H.~Dibekliou\u{g}lu, O.~\c{C}eliktutan,
  B.~G\"{o}kberk, B.~Sankur, and L.~Akarun.
\newblock Bosphorus database for 3{D} face analysis.
\newblock In {\em European Workshop on Biometrics and Identity Management},
  pages 47--56, 2008.

\bibitem{segundo10}
M.~Segundo, L.~Silva, O.~Pereira, and C.~Queirolo.
\newblock Automatic face segmentation and facial landmark detection in range
  images.
\newblock {\em IEEE Transactions on Systems, Man, and Cybernetics, Part B},
  40(5):1319--1330, 2010.

\bibitem{sun01}
Y.~Sun and M.~Abidi.
\newblock Surface matching by 3{D} point's fingerprint.
\newblock In {\em IEEE International Conference on Computer Vision}, volume~2,
  pages 263--269, 2001.

\bibitem{vanKaick11}
O.~van Kaick, H.~Zhang, G.~Hamarneh, and D.~Cohen-Or.
\newblock A survey on shape correspondence.
\newblock {\em Computer Graphics Forum}, 3(6):1681--1707, 2011.

\bibitem{Weise11}
T.~Weise, S.~Bouaziz, H.~Li, and M.~Pauly.
\newblock Realtime performance-based facial animation.
\newblock {\em ACM Transactions on Graphics (SIGGRAPH)}, 30(4):77:1--77:10,
  2011.

\bibitem{wuhrer2010b}
S.~Wuhrer, Z.~Ben~Azouz, and C.~Shu.
\newblock Semi-automatic prediction of landmarks on human models in varying
  poses.
\newblock In {\em Canadian Conference on Computer and Robot Vision}, pages
  136--142, 2010.

\bibitem{wuhrer2011a}
S.~Wuhrer, C.~Shu, and P.~Xi.
\newblock Landmark-free posture invariant human shape correspondence.
\newblock {\em The Visual Computer}, 27(9):843--852, 2011.

\bibitem{Xi09}
P.~Xi and C.~Shu.
\newblock Consistent parameterization and statistical analysis of human head
  scans.
\newblock {\em The Visual Computer}, 25(9):863--871, 2009.

\bibitem{yedidia03}
J.~Yedidia, W.~Freeman, and Y.~Weiss.
\newblock {\em Understanding Belief Propagation and Its Generalizations}.
\newblock Science \& Technology Books, 2003.

\bibitem{yin06}
L.~Yin, X.~Wei, J.~Wang, Y.~Sun, and M.~Rosato.
\newblock A 3{D} facial expression database for facial behavior research.
\newblock In {\em IEEE International Conference on Automatic Face and Gesture
  Recognition}, pages 211--216, 2006.

\end{thebibliography}

\end{document}